\documentclass{article}
\pdfoutput=1
\usepackage{arxiv}

\usepackage[utf8]{inputenc} 
\usepackage[T1]{fontenc}    
\usepackage{hyperref}       
\usepackage{url}            
\usepackage{booktabs}       
\usepackage{amsfonts}       
\usepackage{nicefrac}       
\usepackage{microtype}      
\usepackage{cleveref}       
\usepackage{lipsum}         
\usepackage{graphicx}
\usepackage{natbib}
\usepackage{doi}
\usepackage{amssymb}
\usepackage{multirow}
\usepackage{float}
\usepackage{graphics}
\usepackage{hyperref}
\usepackage{graphicx}
\usepackage{subfigure}
\hypersetup{hypertex=true,
	colorlinks=false,
}

\title{Few-shot semantic segmentation via mask aggregation}

\date{}

\author{Wei Ao\\
	School of Remote Sensing and\\
	Information Engineering, Wuhan\\
	Univerisity, Wuhan, 430079, China\\
	\texttt{wei\_ao@whu.edu.cn} \\
	\And
	Shunyi Zheng\thanks{corresponding author} \\
	School of Remote Sensing and\\
	Information Engineering, Wuhan\\
	Univerisity, Wuhan, 430079, China\\
	\texttt{syzheng@whu.edu.cn} \\
	\And
	Yan Meng\\
	School of Computer Science,\\
	Wuhan University, Wuhan\\
	430072, China \\
	\texttt{mengyan@whu.edu.cn} \\
}

\renewcommand{\undertitle}{}
\begin{document}
\maketitle

\begin{abstract}
Few-shot semantic segmentation aims to recognize novel classes with only very few labelled data. This challenging task requires mining of the relevant relationships between the query image and the support images. Previous works have typically regarded it as a pixel-wise classification problem. Therefore, various models have been designed to explore the correlation of pixels between the query image and the support images. However, they focus only on pixel-wise correspondence and ignore the overall correlation of objects. In this paper, we introduce a mask-based classification method for addressing this problem. The mask aggregation network (MANet), which is a simple mask classification model, is proposed to simultaneously generate a fixed number of masks and their probabilities of being targets. Then, the final segmentation result is obtained by aggregating all the masks according to their locations. Experiments on both the PASCAL-$5^i$ and COCO-$20^i$ datasets show that our method performs comparably to the state-of-the-art pixel-based methods. This competitive performance demonstrates the potential of mask classification as an alternative baseline method in few-shot semantic segmentation. Our source code will be made available at \href{https://github.com/TinyAway/MANet}{https://github.com/TinyAway/MANet}.
\end{abstract}


\section{Introduction}
Semantic segmentation, which aims to generate dense image predictions by assigning a precise category to each pixel of the processed image, is a fundamental problem in computer vision. In recent years, great breakthroughs have been made due to the rapid development of deep convolutional neural networks (DCNNs) \citep{1998Gradient}. Through end-to-end supervised training, deep learning models can quickly acquire the ability to recognize various kinds of objects. Although methods that are based on DCNNs \citep{2015Fully,ronneberger2015u,badrinarayanan2017segnet,zhao2017pyramid,fu2019dual} can obtain good results, they often rely on a large number of annotated images. The performance of the model will deteriorate when there are not adequate training samples.

\begin{figure}[H]
	\centering
	\includegraphics[width=0.6\textwidth,height=6cm]{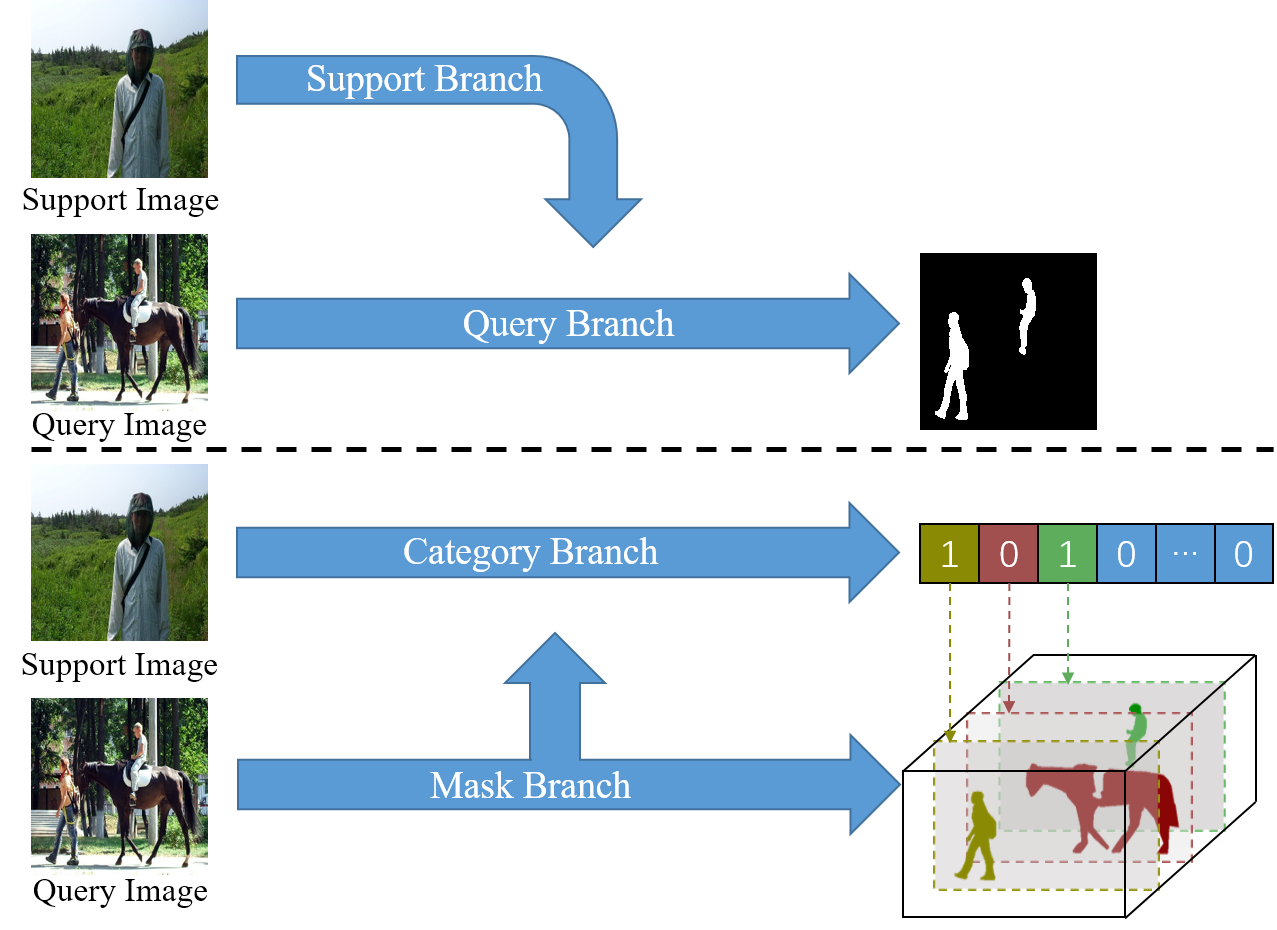}
	\caption{Comparison between the pixel-based few-shot segmentation framework (top) and our proposed mask-based method (bottom). Typical FSS methods focus on classifying all the pixels of the query image through the relevance to the support images. Our method attempts to generate a set of masks and assign them exact categories simultaneously.}
	\label{figure1}
\end{figure}

To address this problem, \citet{shaban2017one} first established a benchmark for few-shot segmentation (FSS). In contrast to regular semantic segmentation methods, which can only segment categories that are included in the training set, FSS methods aim to recognize unseen categories with very few annotated data. The general strategy for tackling the FSS task is to utilize meta-learning. The training data are divided into many episodes, each of which includes one query image and one or a few corresponding annotated support images. Through learning from these episodes, the model can find clues that are related to the query image from the support images and, thus, segment novel classes. Therefore, the key challenge of the FSS task is to exploit information from support images. Recent studies \citep{zhang2019canet,wang2019panet,tian2020prior,yang2020new,liu2020crnet} have generally utilized a two-branch architecture (see Fig. \ref{figure1}) to process both the support images and the query image. The support branch aims to learn prototypes from image-label pairs using masked global average pooling (GAP) \cite{zhou2016learning}, while the query branch is responsible for segmenting the query images by densely comparing the query features and the prototypes. In this way, the recognition ability of the model is no longer limited to the categories that appear in the training set.

Although established methods can achieve good results in FSS, they are all pixel-based, which is not in line with the way human beings perceive objects because humans are better at recognizing whole objects than parts of objects. Therefore, to address this issue, we attempt to explore the association of whole objects between the query image and the support images. Inspired by mask classification methods \citep{he2017mask,hariharan2014simultaneous,kirillov2019panoptic,cheng2021per}, we enable the model to automatically generate a set of masks of all the objects in the query image. Then, the problem of FSS can be transformed into determining whether the masks correspond to the foreground or background. 

In this paper, a mask aggregation network (MANet), which is a simple but effective model, is proposed for tackling the challenges of FSS via mask classification. Similar to the previous FSS methods, we also adopt a two-branch structure, which includes a mask branch and a category branch (see Fig. \ref{figure1}). The mask branch aims to generate a set of binary masks by decoding the query features, while the category branch calculates the probability that each mask is the target by comparing the query features and the support prototypes. All the masks will be aggregated according to their locations to produce the final segmentation result. Finally, to enable the two branches to achieve their goals, a grid loss is designed to guide the model to produce correct masks. Comprehensive experiments on the PASCAL-$5^i$ and COCO-$20^i$ datasets are conducted to evaluate the performance of our method. In summary, the main contributions of this paper are threefold:

\begin{itemize}
\item Based on the perception of whole objects , we develop a novel two-branch model that can solve the problem of FSS by aggregating local mask information. To the best of our knowledge, we are the first to use mask classification for FSS.
\item By leveraging an elaborate grid loss, our model converges well during training and can more accurately extract the masks and the corresponding categories of objects.
\item The proposed method achieves competitive results on the PASCAL-$5^i$ dataset (mIoU: 61.2\% for 1-shot and 64.5\% for 5-shot segmentation) and COCO-$20^i$ dataset (mIoU: 36.4\% for 1-shot and 44.2\% for 5-shot segmentation), which provides compelling evidence that mask classification is an effective method for FSS.	
\end{itemize}

\section{Related work}

\subsection{Semantic segmentation}
Semantic segmentation requires dense prediction of pixels, which is a widely studied problem in the field of computer vision. Mainstream segmentation methods \citep{ronneberger2015u,badrinarayanan2017segnet,chen2018encoder,noh2015learning} typically employ an encoder-decoder structure to realize end-to-end training. The encoder extracts high-dimensional semantic features at low resolution along the convolutional pathways, while the decoder gradually recovers the resolution of the feature maps and obtains a segmentation mask through upsampling blocks. Based on this structure, variants have been proposed for obtaining more accurate results. In U-Net \citep{ronneberger2015u}, skip connections are added between the encoder and the decoder to decrease the loss of scale information in the encoding process. DeepLab V3+ \citep{chen2018encoder} uses several convolution kernels with different scales to accurately capture the spatial information of high-level semantic feature maps. DANet  \citep{fu2019dual} leverages two different attention modules to learn the spatial and channel interdependencies of features. Although these methods have been shown to be effective, they can achieve good results only if many labelled samples are available for training. Moreover, a dramatic decrease in accuracy occurs when they are used to segment unseen classes.

\subsection{Few-shot segmentation}
The objective of few-shot segmentation is to obtain helpful information from few labelled data for segmenting query images that contain novel classes. Recent works often formulate FSS from the perspective of metric learning. PL \citep{dong2018few} produces a prototype for each class via GAP. The segmentation result is obtained by comparing the support prototypes and the query features. In PANet \citep{wang2019panet}, it is assumed that a good FSS model should be able to not only segment the query image through a prototype of the support image but also correctly segment the support image by calculating a prototype of the query image. Therefore, it uses a prototype alignment strategy to fully exploit the support knowledge. CANet \citep{zhang2019canet} uses the convolution operation instead of cosine similarity to adaptively obtain the correlation between the query features and the support features. PFENet \citep{tian2020prior} considers that high-level layers of the encoder can provide rich semantic information. It leverages an extra prior map that is generated from high-level semantic information to further guide the subsequent segmentation steps. The methods that are discussed above are all pixel-based, which obtain segmentation results by comparing the features of the pixels in the query image and the features of the target pixels in the support image. However, they are unstable and easily make mistakes when the query image and the support image differ substantially.

\subsection{Mask classification}
Mask classification is usually used in instance segmentation tasks. To segment all the individual objects, a DCNN is usually employed to automatically generate a set of masks, which are much more numerous than the objects. Mask R-CNN \citep{he2017mask} utilizes a region proposal network (RPN) to generate bounding boxes and then extracts foreground areas from them using an FCN. SOLO \citep{wang2020solo} divides the image into regions, each of which generates a binary mask for the object in it. DETR \citep{carion2020end} further adopts a small fixed number of learnable positional embeddings to transform an image sequence into a mask sequence. However, the accuracies of these methods are limited due to their dependence on object bounding boxes. Recently, MaskFormer \citep{cheng2021per} was proposed, which attempts to represent all the pixels of a category as a mask; this enables mask classification to be used in semantic segmentation. Following this strategy, our model automatically generates masks of all the objects and then divides each mask into foreground or background by making dense comparisons on the query features and the support prototypes.

\section{Problem setting}
The objective of few-shot segmentation is to train a model that can segment unseen classes with only one or a few annotated data. Suppose that the model is trained on a dataset $D_{train}$  with class set $C_{train}$; the goal is to evaluate the model on another dataset $D_{test}$ with class set $C_{test}$, where $C_{train}\cap C_{test}=\varnothing$. In contrast to the common training mode of semantic segmentation, which directly inputs the training samples into the model, in our method, both the training set and the testing set are divided into many episodes. Each episode consists of a query set $Q=(I^q,M^q)$ and a support set $S={(I_i^s,M_i^s )}_{i=1}^K$, where $I^*$ and $M^*$ are the image and its corresponding mask, respectively, and K denotes the number of support images. During training, the model iteratively and randomly samples episodes from $D_{train}$ to learn a mapping from the support set S and query image $I^q$ to the query mask $M^q$. Then, the trained model is evaluated on the testing set in the same way.

\begin{figure}
	\centering
	\includegraphics[width=\textwidth,height=0.3\textheight]{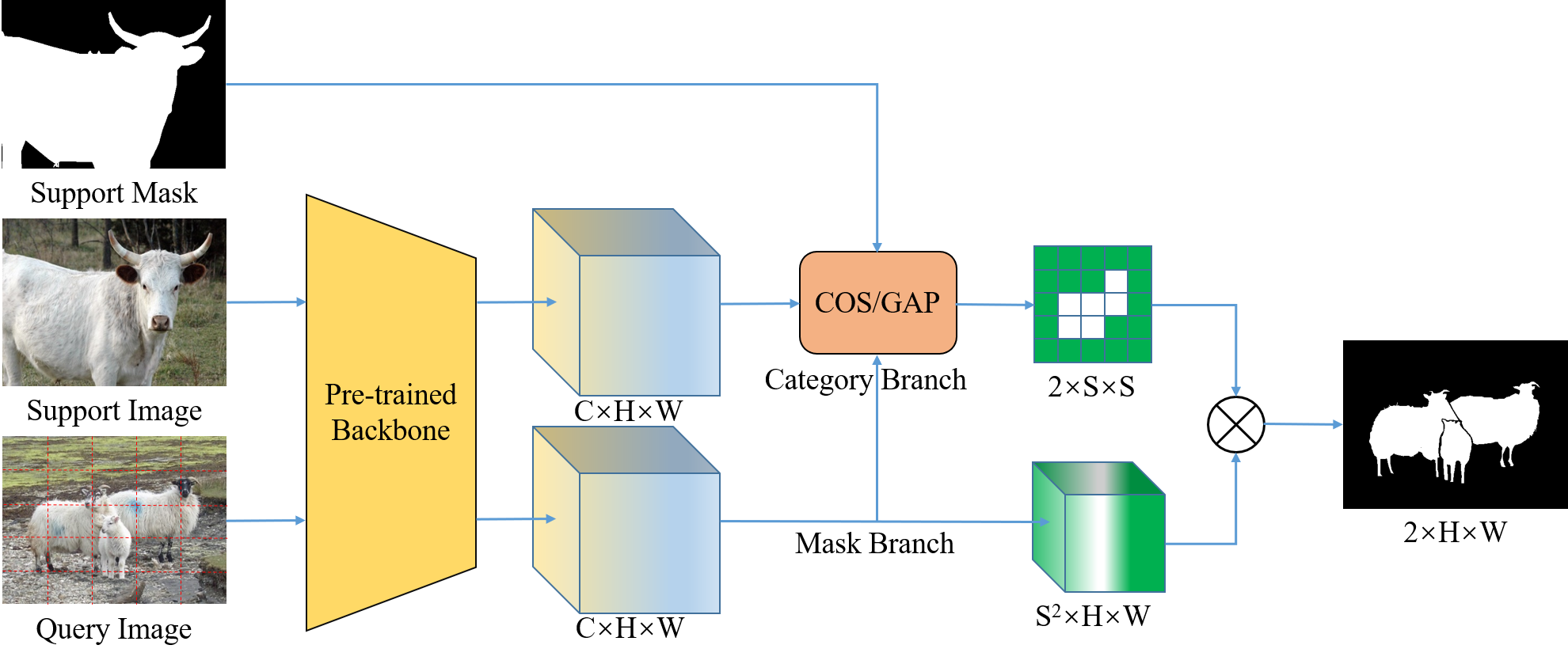}
	\caption{ Overview of the proposed MANet. The query image is divided into an S×S grid. The query image and the support image are first input into a pre-trained backbone network to obtain the query features and the support features, respectively. The cosine similarity of high-level features is calculated to obtain correlation maps, and masked GAP is employed on middle-level features to generate prototypes. Then, the model is divided into two branches. The category branch aims to predict the category of the objects in each grid cell. The mask branch generates a total of $S^2$ masks for all the grid cells. Finally, the segmentation result is produced by matrix multiplication of the categories and the masks.}
	\label{figure2}
\end{figure}

\section{Our method}
In this section, we introduce the architecture of the proposed method and its loss function. Fig. \ref{figure2} illustrates the structure of our proposed MANet. We adopt an encoder-decoder architecture, as is used in most FSS methods. A well-trained encoder (ResNet \citep{he2016deep} pre-trained on ImageNet \citep{krizhevsky2012imagenet}) is used to extract high-dimensional features of the query and support images, and then a decoder is leveraged to predict the final mask by computing the correlation between the query features and the support features. However, in contrast to other FSS methods, as discussed above, we employ a mask classification method to obtain the final prediction. Herein, inspired by YOLO \citep{redmon2016you}, each query image is divided into an S×S grid. If the centre of an object falls within a grid cell, then this grid cell is responsible for extracting the mask and the category of this object. We introduce a two-branch structure for implementing both tasks, which includes a category branch and a mask branch. The purpose of the category branch is to determine whether the object in each grid cell corresponds to the foreground or background, and the mask branch is used to obtain a binary mask of the object in each grid cell. Finally, all the masks are aggregated according to their location information via simple matrix multiplication. Moreover, a grid loss is designed to enable the two branches to perform their respective functions. Our model is presented in detail in the following.

\subsection{Category branch}
\begin{figure}
	\centering
	\includegraphics[width=0.6\textwidth,height=7cm]{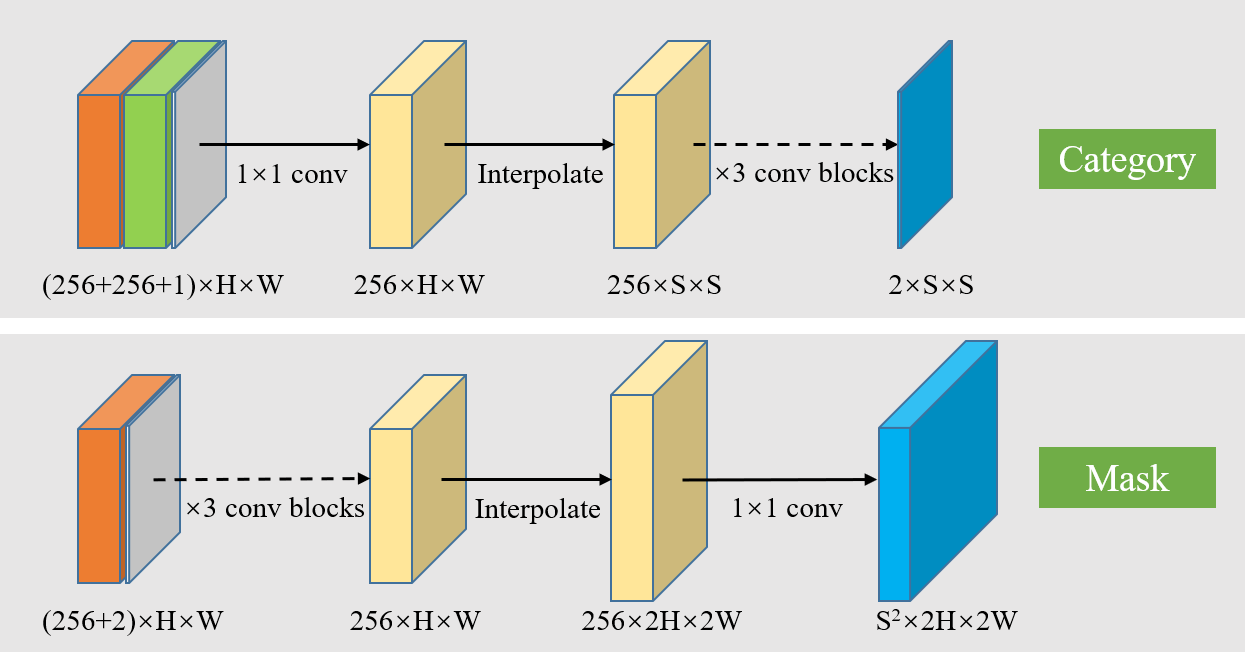}
	\caption{Architectures of the category branch and the mask branch. Each branch is mainly composed of three convolutional blocks. The category branch converts three inputs (the query feature maps, the expanded prototype and the normalized correlation map) into categories that correspond to each grid cell. The mask branch obtains the masks of the objects in each grid cell by combining the query features and the spatial coordinate information.}
	\label{figure3}
\end{figure}

The category branch is used to obtain the categories to which the objects in each grid cell belong. This branch has three inputs: query feature maps, an expanded prototype and a correlation map. The query feature maps are the middle-level semantic features that are extracted from the query image by the encoder. The dimension of each feature map is compressed to 256 via a 1×1 convolution operation. The prototype is obtained by using masked GAP for middle-level support features. Then, it is expanded to the same size as the query feature maps. The correlation map is similar to the prior mask that is used in PFENet \citep{tian2020prior}. First, the cosine similarity $C_Q$ is calculated between high-level query features $x_q\in x_Q$ and support features $x_s\in x_S$ as
\begin{equation}
	\cos \left( {{x_q},{x_s}} \right) = \frac{{x_q^T{x_s}}}{{\left\| {{x_q}} \right\|\left\| {{x_s}} \right\|}},q,s \in \left\{ {1,2,...,HW} \right\}
	\label{eq1}
\end{equation}
where H and W are the height and width, respectively, of the query feature maps. Then, $C_Q$ is normalized to between 0 and 1 as
\begin{equation}
	{C_Q} = \frac{{{C_Q} - \min \left( {{C_Q}} \right)}}{{\max \left( {{C_Q}} \right) - \min \left( {{C_Q}} \right) + \varepsilon }}
	\label{eq2}
\end{equation}
where $\varepsilon$ is a constant, which is set to 1e-7. After all the inputs are obtained, they are concatenated along the channel dimension. Thus, the channel dimension becomes 256+256+1. Then, we compress it to 256 dimensions with a 1×1 convolution operation. As indicated in Fig. \ref{figure3}, this branch consists of three convolutional blocks. The integrated feature maps are first interpolated to size S×S; then, the probability that each grid cell represents the foreground or background is output through the convolutional blocks.

In the few-shot setting, we conduct the same process with PFENet \citep{tian2020prior}. The support prototypes and correlation maps that are produced by different support features are simply replaced by their mean values. Using this simple process, we achieve an obvious accuracy improvement compared to 1-shot segmentation without increasing the number of parameters of the model (see Table \ref{Table 1}).

\subsection{Mask branch}
As discussed above, we divide each query image into an S×S grid. The purpose of the mask branch is to predict a binary mask for each grid cell. We stack all the masks together along the channel dimension; hence, the channel dimension of the output will be $S^2$. The $k^{th}$ channel represents the mask of the object in grid cell (i,j), where  $k=i*S+j,i,j\in \{0,1,…,S-1\}$. As shown in Fig. \ref{figure3}, we also use an FCN with three convolutional blocks to achieve this goal. However, the traditional convolution operation is spatially invariant, which is not desirable for our model because we need the model to be sensitive to the spatial information to obtain the masks of the objects that are located in each grid. Therefore, inspired by the CoordConv solution \citep{liu2018intriguing}, we directly splice the x and y coordinate information and the query feature maps together to enable the convolutional filters to know their spatial locations. Thus, the third dimension of the new tensor becomes D+2, where D is the channel dimension of the query feature maps. Moreover, to prevent coordinate values that are too large from affecting the training of the model, they are normalized to [-1,1]. Then, we use the convolutional blocks to extract object features. The output is upsampled to twice the size of the original feature maps to obtain more accurate masks. Finally, a 1×1 convolution operation is employed to generate $S^2$  masks.

\subsection{Loss function}
The loss function that we use consists of two parts: a pixel loss and a grid loss. The pixel loss is a cross-entropy loss that is calculated between the final prediction and the ground truth. It is defined as
\begin{equation}
	{{\rm{L}}_{pixel}} =  - \frac{1}{N}\sum\nolimits_{i = 1}^N {{y_i}} \log {p_i}
	\label{eq3}
\end{equation}
where $p_i$ and $y_i$ denote the prediction and label, respectively, of pixel i and N is the total number of pixels.

However, we observe that only calculating the pixel loss cannot result in satisfactory performance (see Table \ref{Table 1}). This is because in this approach, there are no additional constraints on the two branches, and it is essentially the same as the common pixel-wise classification methods. Therefore, to ensure that the two branches can play their respective roles, we calculate a grid loss for the category branch. We assume that the more pixels there are of the target object in a grid cell, the greater the probability that the mask that corresponds to this grid cell belongs to the foreground. For each query mask, we first calculate the mean value of each grid $G_m$ and then normalize it to [0,1] as follows:
\begin{equation}
	{G_m} = \frac{{{G_m} - \min \left( {{G_m}} \right)}}{{\max \left( {{G_m}} \right) - \min \left( {{G_m}} \right) + \varepsilon }}
	\label{eq4}
\end{equation}
where $\varepsilon$ is set to the same value as Eq. (\ref{eq2}). The grid loss $L_{grid}$ is defined as
\begin{equation}
	{L_{grid}} =  - \frac{1}{{{S^2}}}\sum\nolimits_{i = 1}^{{S^2}} {{G_{m,i}}} \log \left( {soft\max \left( {{g_i}} \right)} \right)
	\label{eq5}
\end{equation}
where $g_i$ represents the category of the $i^{th}$ grid cell that is predicted by the category branch. Thus, the final loss function is expressed as
\begin{equation}
	L = {L_{pixel}} + \lambda {L_{grid}}
	\label{eq6}
\end{equation}
where $\lambda$ is a constant and is set to 1 in our experiments.

\section{Experiments}
\subsection{Datasets and evaluation metrics}
Two FSS benchmarks, namely, PASCAL-$5^i$ and COCO-$20^i$, are selected to evaluate our method. PASCAL-$5^i$ is composed of PASCAL VOC 2012 \cite{everingham2010pascal} and extra annotations from the SDS \cite{hariharan2014simultaneous} dataset. It contains a total of 20 object categories and is evenly divided into four folds. For each fold, 15 classes are used for training, and the remaining 5 classes are used for testing. The COCO-$20^i$ dataset, which is built from MS-COCO \citep{lin2014microsoft}, is a more challenging dataset that consists of 80 object categories. Likewise, all the classes are split into 4 folds. Thus, each fold has 20 classes. For each fold, 60 base classes are used for training, and the remaining 20 classes are used for testing.

Following the work of \citet{shaban2017one}, the mean intersection over union (mIoU) is adopted as our main evaluation metric. It is calculated by averaging the IoUs of all the categories. Moreover, FB-IoU is also computed, which can measure the model's ability to distinguish the foreground and the background. For PASCAL-$5^i$, we randomly sample 1,000 episodes and calculate the average of the five runs as the final mIoU/FB-IoU. For COCO-$20^i$, because of its large size, 20,000 episodes are sampled randomly to obtain stable results.

\subsection{Implementation details}
Our model is implemented with PyTorch on a single RTX 3080. We employ ResNet-50 and ResNet-101 as the backbones, which are pre-trained on ImageNet \citep{krizhevsky2012imagenet}. The parameters of the pre-trained backbone networks are frozen when the model is under training. Adam is leveraged to optimize the model, and the learning rate is set to 1e-4. All the images are resized to 473×473 pixels before they are fed to the model. During inference, the size of the output mask is recovered to the original size of the corresponding query images. Several data augmentation strategies, such as horizontal flipping, random scaling, random rotation and random shifting, are utilized. We set the batch size to 4 and train our model for 100 and 20 epochs on PASCAL-$5^i$ and COCO-$20^i$, respectively. Notably, we directly output the final prediction result without fine-tuning or any other post-processing methods (e.g., DenseCRF \citep{chen2017deeplab}).
\subsection{Results and analysis}
Table \ref{Table 1} presents the mIoU and FB-IoU values that are obtained by our model on PASCAL-$5^i$. We build our model on two backbones, namely, ResNet-50 and ResNet-101, and compare it with various state-of-the-art methods. When the model is built on ResNet-101, the mIoU values of the 1-shot and 5-shot experiments are 0.4\% and 3.7\% higher, respectively, than those on PFENet \citep{tian2020prior}. Notably, our model does not use multi-scale features and has only 5.5 M trainable parameters, but it can outperform the compared methods. In terms of FB-IoU, our model outperforms PFENet  regardless of whether it uses ResNet -50 or ResNet -101 as the backbone in the 5-shot experiments. We also report the performance without using the grid loss. When the grid loss is not used in the training process, the performances in the 1-shot and 5-shot experiments severely deteriorate. In addition, the results of 5-shot segmentation are no longer significantly improved compared to those of 1-shot segmentation. This is probably because the model is not guided to extract the correct mask and it degenerates into a per-pixel classification model.

\begin{table}
	\caption{mIoU and FB-IoU of 1-shot and 5-shot segmentation on the PASCAL-$5^i$ dataset using the proposed method and state-of-the-art few-shot segmentation methods. Params represents the number of learnable parameters. † indicates that the grid loss is not used during training. The best performances and the minimum parameters are highlighted in bold.}
	\resizebox{\textwidth}{!}
	{\begin{tabular}{c|c|cccccc|cccccc|c}
			\hline
			\multirow{2}{*}{Backbone}   & \multirow{2}{*}{Method} & \multicolumn{6}{c|}{1-shot}                                                                                        & \multicolumn{6}{c|}{5-shot}                                                                                        & \multirow{2}{*}{Params} \\ \cline{3-14}
			&                         & Fold0         & Fold1         & Fold2         & Fold3         & \multicolumn{1}{c|}{Mean}          & FB-IoU       & Fold0         & Fold1         & Fold2         & Fold3         & \multicolumn{1}{c|}{Mean}          & FB-IoU       &                         \\ \hline
			\multirow{6}{*}{ResNet-50}  & PANet \cite{wang2019panet}            & 44.0          & 57.5          & 50.8          & 44.4          & \multicolumn{1}{c|}{49.1}          & 66.5          & 55.3          & 67.2          & 61.3          & 53.2          & \multicolumn{1}{c|}{59.3}          & 70.7          & 14.7 M                   \\
			& CANet \citep{zhang2019canet}            & 52.5          & 65.9          & 51.3          & 51.9          & \multicolumn{1}{c|}{55.4}          & 66.2          & 55.5          & 67.8          & 51.9          & 53.2          & \multicolumn{1}{c|}{57.1}          & 69.6          & 19.0 M                   \\
			& PFENet \citep{tian2020prior}        & 61.7          & 69.5          & 55.4          & 56.3          & \multicolumn{1}{c|}{60.8}          & \textbf{72.0} & 63.1          & 70.7          & 55.8          & 57.9          & \multicolumn{1}{c|}{61.9}          & 72.3          & 10.4 M                   \\
			& RePRI \citep{boudiaf2021few}           & 60.2          & 67.0          & \textbf{61.7} & 47.5          & \multicolumn{1}{c|}{59.1}          & -             & 64.5          & 70.8          & \textbf{71.7} & 60.3          & \multicolumn{1}{c|}{\textbf{66.8}} & -             & -                       \\
			& SCL \citep{zhang2021self}           & \textbf{63.0} & \textbf{70.0} & 56.5          & 57.7          & \multicolumn{1}{c|}{\textbf{61.8}} & 71.9          & 64.5          & 70.9          & 57.3          & 58.7          & \multicolumn{1}{c|}{62.9}          & \textbf{72.8} & -                       \\ \cline{2-15} 
			& MANet(Ours)             & 62.0          & 69.4          & 51.8          & \textbf{58.2} & \multicolumn{1}{c|}{60.3}          & 71.4          & \textbf{66.0} & \textbf{71.6} & 55.1          & \textbf{64.5} & \multicolumn{1}{c|}{64.3}          & \textbf{75.2} & \textbf{5.5 M}           \\ \hline
			\multirow{7}{*}{ResNet-101} & PPNet \citep{liu2020part}           & 52.7          & 62.8          & 57.4          & 47.7          & \multicolumn{1}{c|}{55.2}          & 70.9          & 60.3          & 70.0          & \textbf{69.4} & 40.7          & \multicolumn{1}{c|}{65.1}          & \textbf{77.5} & 50.5 M                   \\
			& FWB \citep{nguyen2019feature}           & 51.3          & 64.5          & 56.7          & 52.2          & \multicolumn{1}{c|}{56.2}          & -             & 54.8          & 67.4          & 62.2          & 55.3          & \multicolumn{1}{c|}{59.9}          & -             & 43.0 M                  \\
			& DAN \citep{wang2020few}            & 54.7          & 68.6          & 57.8          & 51.6          & \multicolumn{1}{c|}{58.2}          & 71.9          & 57.9          & 69.0          & 60.1          & 54.9          & \multicolumn{1}{c|}{60.5}          & 72.3          & -                       \\
			& PFENet \citep{tian2020prior}          & 61.7          & 69.5          & 55.4          & 56.3          & \multicolumn{1}{c|}{60.8}          & \textbf{72.9} & 63.1          & 70.7          & 55.8          & 57.9          & \multicolumn{1}{c|}{61.9}          & 73.5          & 10.8 M                   \\
			& RePRI \citep{boudiaf2021few}           & 56.9          & 68.6          & \textbf{62.2} & 47.2          & \multicolumn{1}{c|}{59.4}          & -             & 66.2          & \textbf{71.4} & 67.0          & 57.7          & \multicolumn{1}{c|}{\textbf{65.6}} & -             & -                       \\ \cline{2-15} 
			& MANet†(Ours)             & \textbf{64.0} & 68.1          & 52.2          & 57.0          & \multicolumn{1}{c|}{60.3}          & 70.9          & 64.1          & 69.2          & 52.3          & 56.5          & \multicolumn{1}{c|}{60.5}          & 71.0          & \textbf{5.5 M}           \\ \cline{2-15} 
			& MANet(Ours)             & 63.9          & \textbf{69.2} & 52.5          & \textbf{59.1} & \multicolumn{1}{c|}{\textbf{61.2}} & 71.4          & \textbf{67.0} & 70.8          & 54.8          & \textbf{65.5} & \multicolumn{1}{c|}{64.5}          & 74.1          & \textbf{5.5 M}           \\ \hline
	\end{tabular}}
	\label{Table 1}
\end{table}

\begin{table}
	\caption{mIoU of 1-shot and 5-shot segmentation on the COCO-$20^i$ dataset using the proposed method and state-of-the-art few-shot segmentation methods. The best performances are highlighted in bold.}
	\resizebox{\textwidth}{!}
	{\begin{tabular}{c|c|ccccc|ccccc}
			\hline
			\multirow{2}{*}{Backbone}  & \multirow{2}{*}{Method} & \multicolumn{5}{c|}{1-shot}                                  & \multicolumn{5}{c}{5-shot}                                                    \\ \cline{3-12} 
			&                         & Fold1         & Fold2         & Fold3         & Fold4 & Mean & Fold1         & Fold2         & Fold3         & Fold4         & Mean          \\ \hline
			\multirow{6}{*}{ResNet-50} & FWB \cite{nguyen2019feature}             & 19.9          & 18.0          & 21.0          & 28.9  & 21.2 & 19.1          & 21.5          & 23.9          & 30.7          & 23.7          \\
			& PRMM \citep{yang2020prototype}           & 29.5          & 36.8          & 28.9          & 27.0  & 30.6 & 33.8          & 42.0          & 33.0          & 33.3          & 35.5          \\
			& PPNet \citep{liu2020part}           & 28.1          & 30.8          & 29.5          & 27.7  & 29.0 & 39.0          & 40.8          & 37.1          & 37.3          & 38.5          \\
			& PFENet \citep{tian2020prior}          & \textbf{34.3} & 33.0          & 32.3          & 30.1  & 32.4 & 38.5          & 38.6          & 38.2          & 34.3          & 37.4          \\
			& RePRI \citep{boudiaf2021few}           & 31.2          & 38.1          & 33.3          & 33.0  & 34.0 & 38.5          & 46.2          & 40.0          & \textbf{43.6} & 42.1          \\ \cline{2-12} 
			& MANet(Ours)             & 33.9          & \textbf{40.6} & \textbf{35.7} & \textbf{35.2}  & \textbf{36.4} & \textbf{41.9} & \textbf{49.1} & \textbf{43.2} & 42.7          & \textbf{44.2} \\ \hline
		\end{tabular}
	}
	\label{Table 2}
	
\end{table}

The experimental results on COCO-$20^i$ are presented in Table \ref{Table 2}. It is obvious that our MANet achieves a higher mIoU than the state-of-the-art methods. It achieves 4.0\% (1-shot) and 6.8\% (5-shot) mIoU improvements over PFENet \citep{tian2020prior} using the ResNet-50 backbone. Due to the wide variety and the large differences in object scales of this dataset, 1-shot segmentation fails when the support features are ambiguous. However, when 5-shot segmentation is performed, the model obtains the category of each mask more accurately by averaging all the support features. Therefore, in the 5-shot experiment, a 7.8\% mIoU improvement over the 1-shot experiment is realized.

\begin{figure}
	\centering
	
	\subfigure{
		\begin{minipage}[t]{3mm}
			\rotatebox{90}{~~~~~~~~~~\rotatebox{-90}{\scriptsize{(a)}}}
		\end{minipage}

		\begin{minipage}[t]{0.13\linewidth}
			\centering
			\includegraphics[width=1\linewidth]{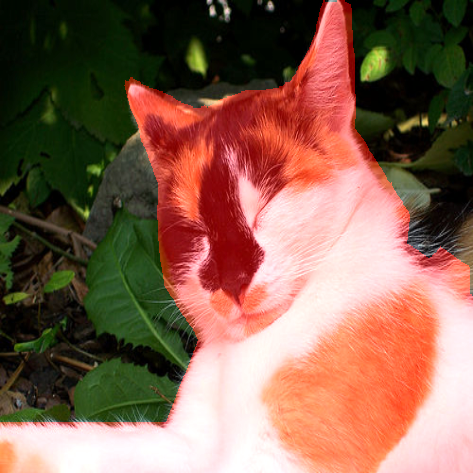}
		\end{minipage}
		\begin{minipage}[t]{0.13\linewidth}
			\centering
			\includegraphics[width=1\linewidth]{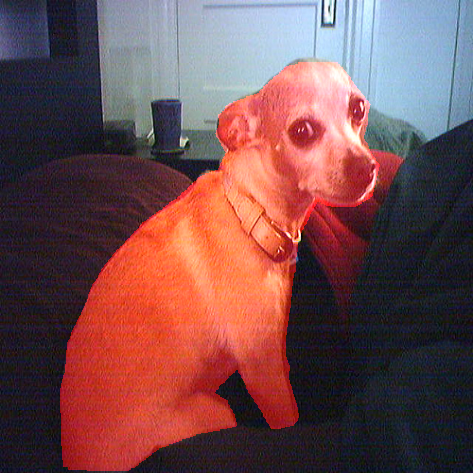}
		\end{minipage}
		\begin{minipage}[t]{0.13\linewidth}
			\centering
			\includegraphics[width=1\linewidth]{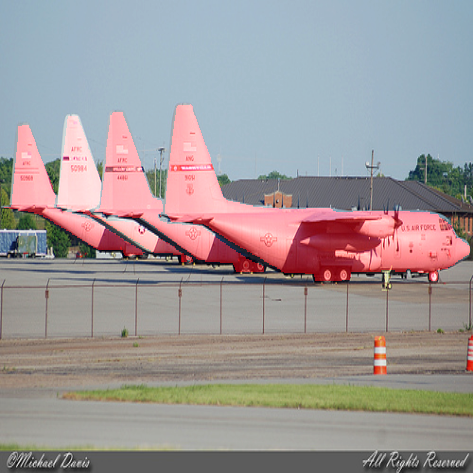}
		\end{minipage}
		\begin{minipage}[t]{0.13\linewidth}
			\centering
			\includegraphics[width=1\linewidth]{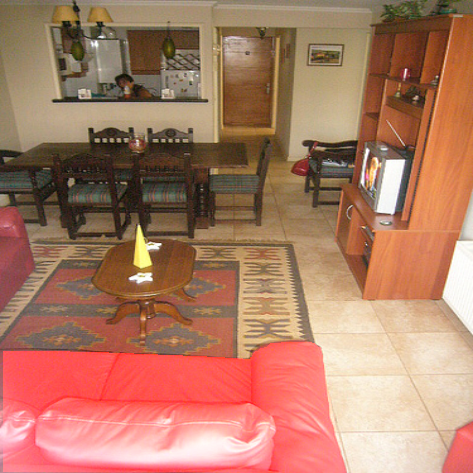}
		\end{minipage}
		\begin{minipage}[t]{0.13\linewidth}
			\centering
			\includegraphics[width=1\linewidth]{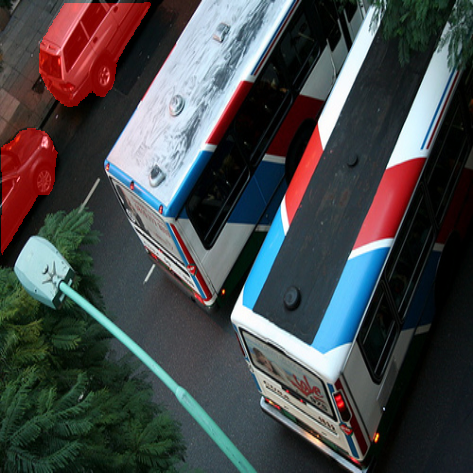}
		\end{minipage}
		\begin{minipage}[t]{0.13\linewidth}
			\centering
			\includegraphics[width=1\linewidth]{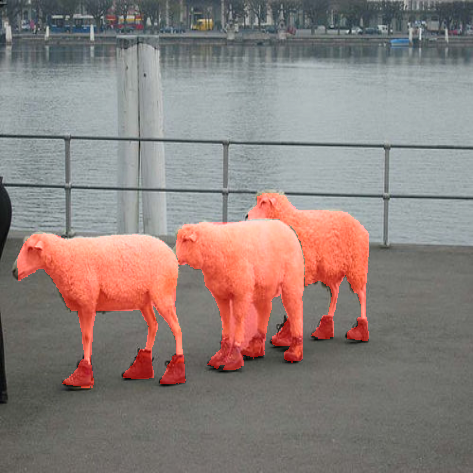}
		\end{minipage}
		\begin{minipage}[t]{0.13\linewidth}
			\centering
			\includegraphics[width=1\linewidth]{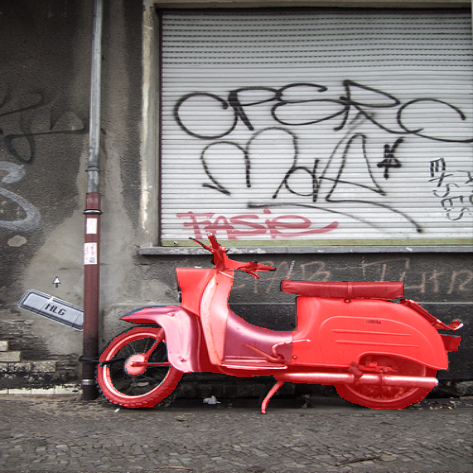}
		\end{minipage}	
	}
	
	\vspace{-3mm}
	
	\subfigure{
		\begin{minipage}[t]{3mm}
			\rotatebox{90}{~~~~~~~~~~\rotatebox{-90}{\scriptsize{(b)}}}
		\end{minipage}
		
		\begin{minipage}[t]{0.13\linewidth}
			\centering
			\includegraphics[width=1\linewidth]{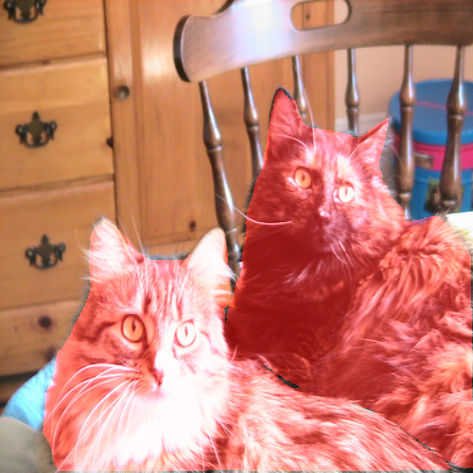}
		\end{minipage}
		\begin{minipage}[t]{0.13\linewidth}
			\centering
			\includegraphics[width=1\linewidth]{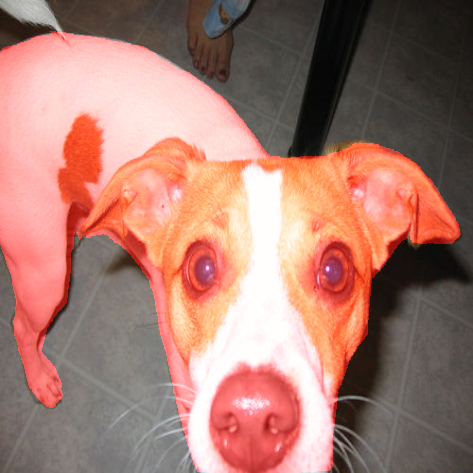}
		\end{minipage}
		\begin{minipage}[t]{0.13\linewidth}
			\centering
			\includegraphics[width=1\linewidth]{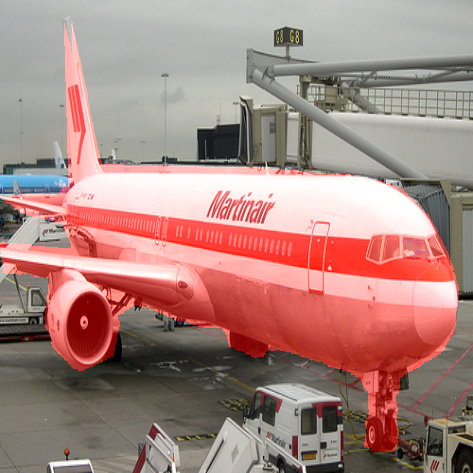}
		\end{minipage}
		\begin{minipage}[t]{0.13\linewidth}
			\centering
			\includegraphics[width=1\linewidth]{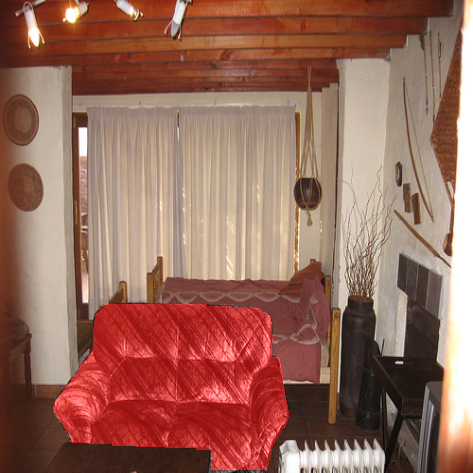}
		\end{minipage}
		\begin{minipage}[t]{0.13\linewidth}
			\centering
			\includegraphics[width=1\linewidth]{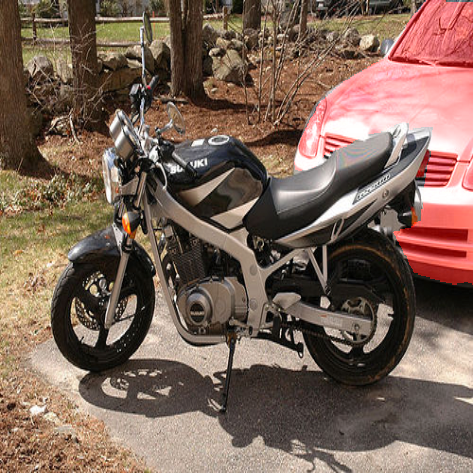}
		\end{minipage}
		\begin{minipage}[t]{0.13\linewidth}
			\centering
			\includegraphics[width=1\linewidth]{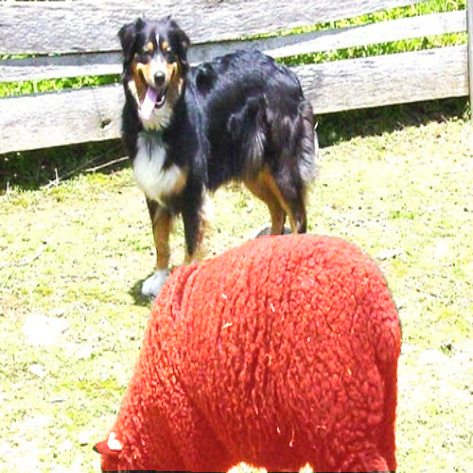}
		\end{minipage}
		\begin{minipage}[t]{0.13\linewidth}
			\centering
			\includegraphics[width=1\linewidth]{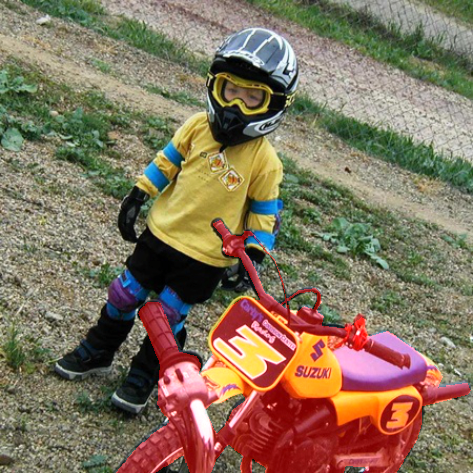}
		\end{minipage}	
	}
	
	\vspace{-3mm}
	
	\subfigure{
		\begin{minipage}[t]{3mm}
			\rotatebox{90}{~~~~~~~~~~\rotatebox{-90}{\scriptsize{(c)}}}
		\end{minipage}
		
		\begin{minipage}[t]{0.13\linewidth}
			\centering
			\includegraphics[width=1\linewidth]{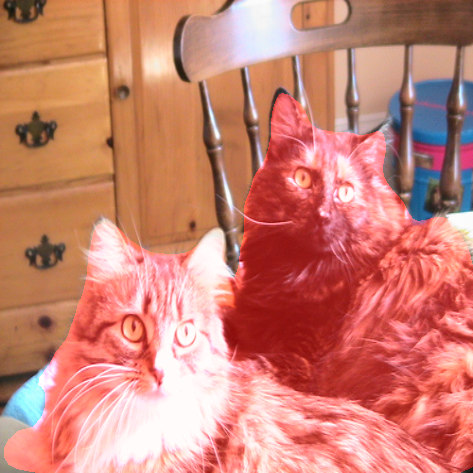}
		\end{minipage}
		\begin{minipage}[t]{0.13\linewidth}
			\centering
			\includegraphics[width=1\linewidth]{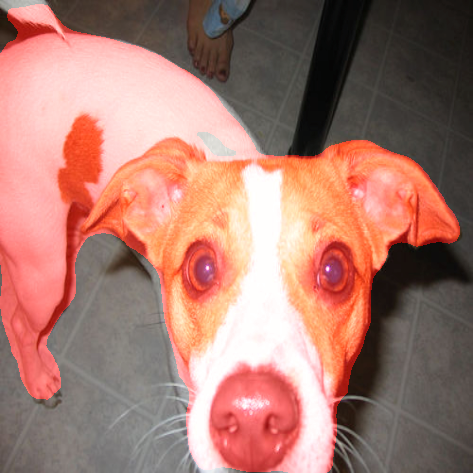}
		\end{minipage}
		\begin{minipage}[t]{0.13\linewidth}
			\centering
			\includegraphics[width=1\linewidth]{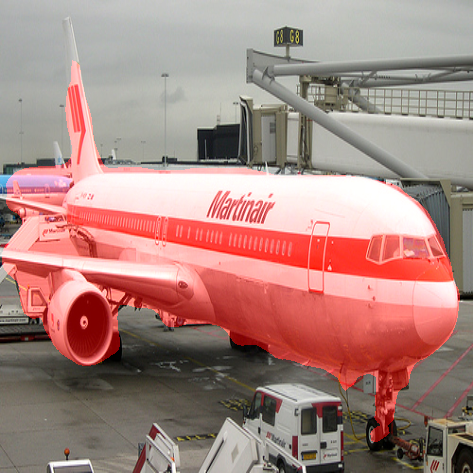}
		\end{minipage}
		\begin{minipage}[t]{0.13\linewidth}
			\centering
			\includegraphics[width=1\linewidth]{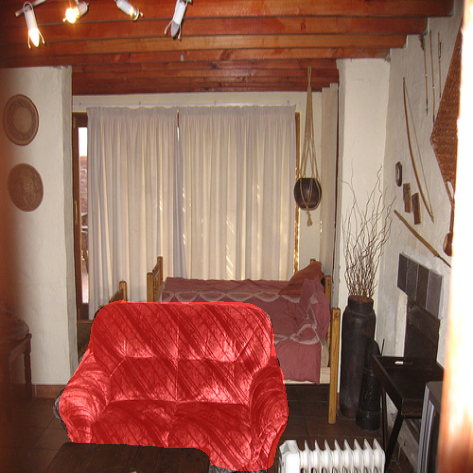}
		\end{minipage}
		\begin{minipage}[t]{0.13\linewidth}
			\centering
			\includegraphics[width=1\linewidth]{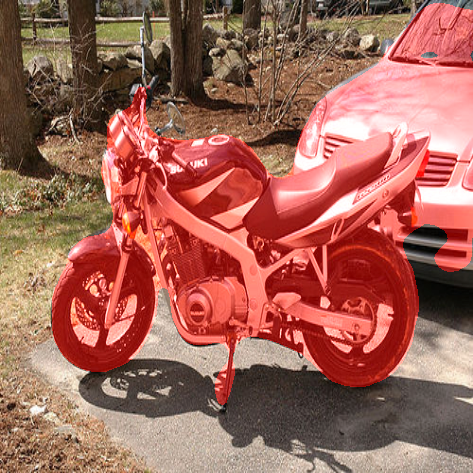}
		\end{minipage}
		\begin{minipage}[t]{0.13\linewidth}
			\centering
			\includegraphics[width=1\linewidth]{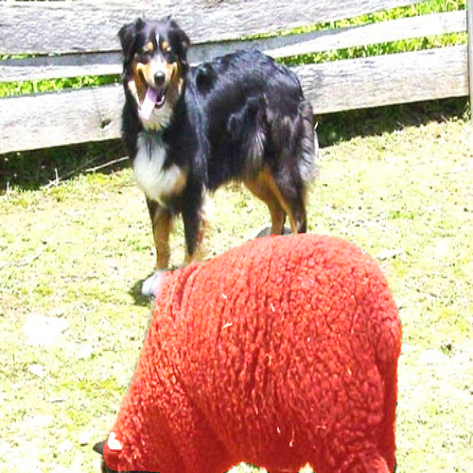}
		\end{minipage}
		\begin{minipage}[t]{0.13\linewidth}
			\centering
			\includegraphics[width=1\linewidth]{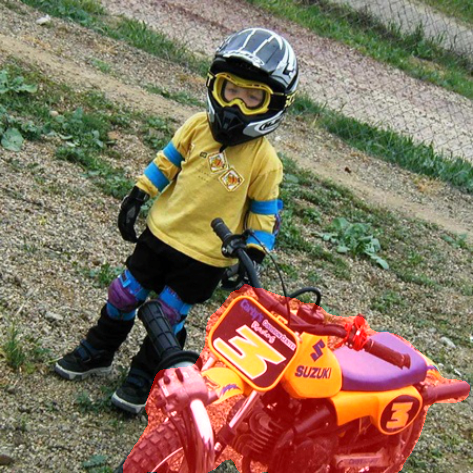}
		\end{minipage}	
	}
	
	\vspace{-3mm}
	
	\subfigure{
		\begin{minipage}[t]{3mm}
			\rotatebox{90}{~~~~~~~~~~\rotatebox{-90}{\scriptsize{(d)}}}
		\end{minipage}
		
		\begin{minipage}[t]{0.13\linewidth}
			\centering
			\includegraphics[width=1\linewidth]{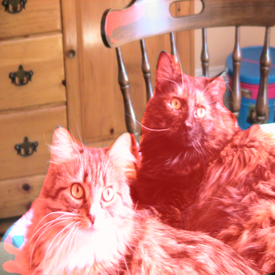}
		\end{minipage}
		\begin{minipage}[t]{0.13\linewidth}
			\centering
			\includegraphics[width=1\linewidth]{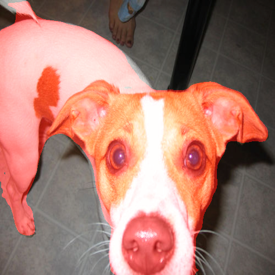}
		\end{minipage}
		\begin{minipage}[t]{0.13\linewidth}
			\centering
			\includegraphics[width=1\linewidth]{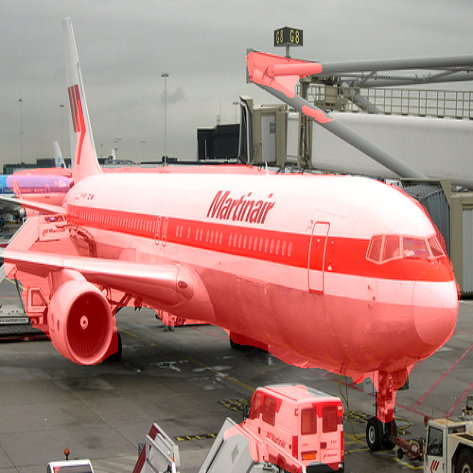}
		\end{minipage}
		\begin{minipage}[t]{0.13\linewidth}
			\centering
			\includegraphics[width=1\linewidth]{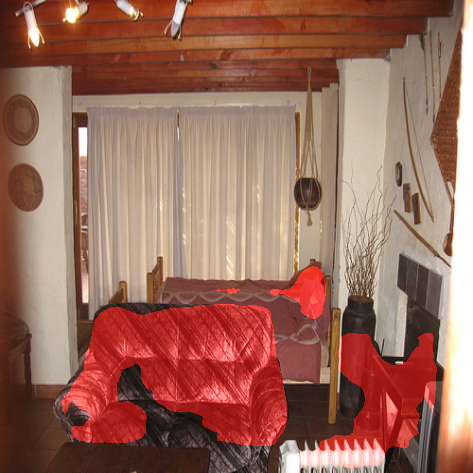}
		\end{minipage}
		\begin{minipage}[t]{0.13\linewidth}
			\centering
			\includegraphics[width=1\linewidth]{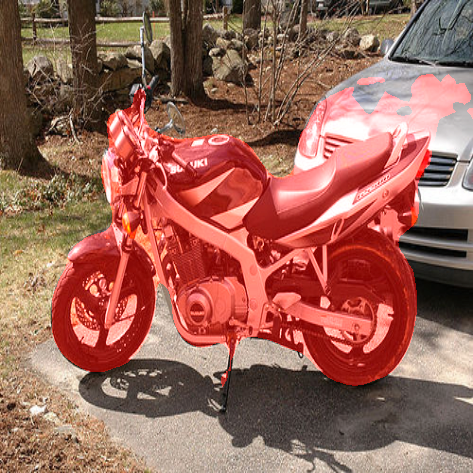}
		\end{minipage}
		\begin{minipage}[t]{0.13\linewidth}
			\centering
			\includegraphics[width=1\linewidth]{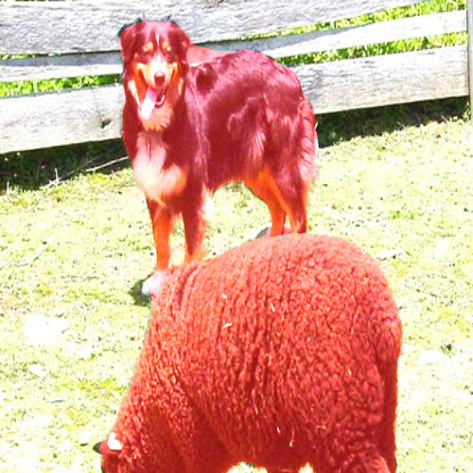}
		\end{minipage}
		\begin{minipage}[t]{0.13\linewidth}
			\centering
			\includegraphics[width=1\linewidth]{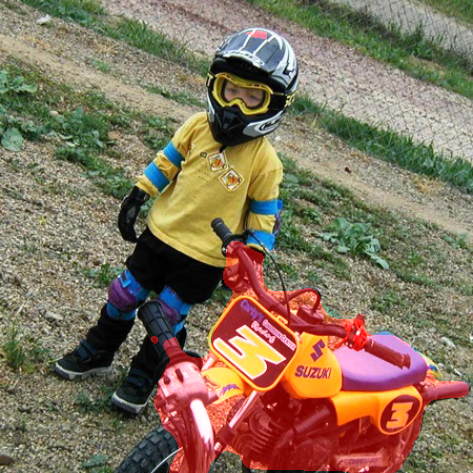}
		\end{minipage}	
	}

	\caption{Qualitative results on our MANet on PASCAL-$5^i$  dataset. From top to bottom: (a) Support images for the 1-shot task and their masks; (b) the query images and their masks, (c) the predictions of MANet, and (d) the predictions of MANet without the grid loss.}
	
	\label{figure4}
\end{figure}

Fig. \ref{figure4} shows several qualitative results of our model. The evaluation samples are from PASCAL-$5^i$ We visualize and compare the results of MANet with the case in which the grid loss is not used. We observe that the masks that are generated by the model with the grid loss are more complete. Moreover, it is obvious that the grid loss enables the model to classify the masks more accurately. In contrast, MANet fails to obtain accurate object boundaries when the grid loss is not utilized. This is probably because when there is no restriction on the content that is generated by each grid cell, excessive training causes network overfitting. Thus, it gradually becomes a per-pixel classifier, which is vulnerable to boundary noise.

\subsection{Ablation study of grid size}
The proposed MANet aims to generate a fixed number of masks and integrate them according to their spatial locations. Therefore, the grid size is an important parameter of the model. In this section, we study the influence of it on the model. We conduct an ablation study on PASCAL-$5^i$ using ResNet-50 as the backbone. We divide the query images into grids of various sizes (12×12, 24×24, and 36×36). Both 1-shot and 5-shot experiments are conducted for comparison. Table \ref{Table 3} presents the mIoU and FB-IoU values for various grid sizes in the proposed MANet. We observe that the performance of the model is not greatly affected by the grid size. The result that is obtained using a 12×12 grid is equivalent to that obtained using a 36×36 grid but requires fewer calculations. This is probably because there are fewer small objects in the PASCAL-$5^i$ dataset and a 12×12 grid is sufficient for extracting most objects.

\begin{table}[H]
	\centering
	\caption{Ablation study of grid size on PASCAL-$5^i$}
	\begin{tabular}{c|cc|cc}
		\hline
		\multirow{2}{*}{Grid Number} & \multicolumn{2}{c|}{1-shot} & \multicolumn{2}{c}{5-shot} \\ \cline{2-5} 
		& mIoU        & FB-IoU        & mIoU        & FB-IoU       \\ \hline
		12×12                        & 60.3        & 71.4          & 64.3        & 75.2         \\
		24×24                        & 59.9        & 71.4          & 64.6        & 75.2         \\
		36×36                        & 59.3        & 71.1          & 64.7        & 75.0         \\ \hline
	\end{tabular}
	\label{Table 3}
\end{table}

\subsection{Visualization of masks}
\begin{figure}
	\centering
	\includegraphics[width=\textwidth,height=8cm]{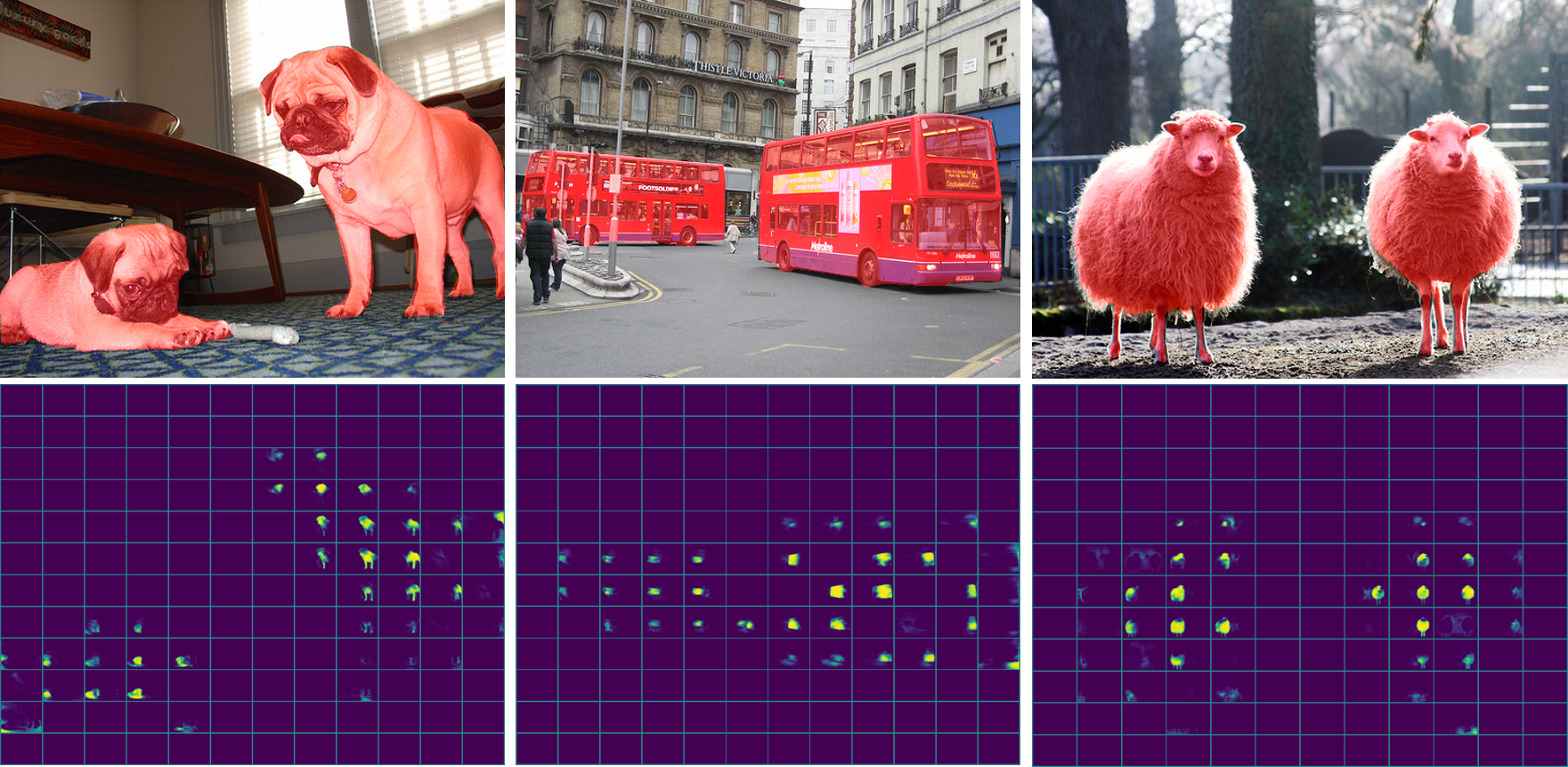}
	\caption{ Visualization of generated masks. The top row presents query images and corresponding segmentation results. The bottom row shows all the masks that are produced by the mask branch of the model. The grid size is 12×12; thus, the total number of masks is 144. Each subfigure represents the mask that is generated at the corresponding location. To facilitate observation, we only show the foreground masks that affect the segmentation results. }
	\label{figure5}
\end{figure}

To better understand why our method is effective, we visualize the concrete masks that are generated by the model with a grid of size 12×12. In Fig. \ref{figure5}, we show all 144 masks of a query image that are produced by the mask branch at the corresponding positions. Subfigure (i,j) is the binary mask of the $k^{th}$ channel dimension of the mask tensor, where $k=i*S+j,i,j\in\{0,1,…,S-1\}$. To clearly present the contents of the masks, we only show the masks that are predicted to correspond to the foreground by the category branch because the masks that are judged as corresponding to the background will not contribute to the segmentation result. We observe that each grid cell generates a mask for the objects that are located in it, and the mask that is produced by each grid cell is highly correlated with its location. Therefore, the final prediction result can be easily obtained by aggregating all the foreground masks.

\section{Conclusions}
We have presented a novel framework for few-shot semantic segmentation that is based on mask classification in this paper. The proposed MANet transforms FSS into a problem of classifying objects in each region by generating a set of binary masks. The segmentation result is obtained by assigning a category to each mask and aggregating all the masks that are predicted to correspond to the foreground. Leveraging a well-designed loss function, MANet can accurately extract the target masks through a simple two-branch structure. Experiments on the PASCAL-$5^i$ and COCO-$20^i$ datasets demonstrate that MANet outperforms the state-of-the-art pixel-based FSS methods. In addition to obtaining highly satisfactory results, it has very few parameters and is easy to combine with other methods. Therefore, we believe that the mask classification approach can be a potential baseline method for few-shot semantic segmentation.

\bibliographystyle{unsrtnat}
\bibliography{mybibfile}

\begin{thebibliography}{35}
\providecommand{\natexlab}[1]{#1}
\providecommand{\url}[1]{\texttt{#1}}
\expandafter\ifx\csname urlstyle\endcsname\relax
  \providecommand{\doi}[1]{doi: #1}\else
  \providecommand{\doi}{doi: \begingroup \urlstyle{rm}\Url}\fi

\bibitem[Lecun and Bottou(1998)]{1998Gradient}
Y.~Lecun and L.~Bottou.
\newblock Gradient-based learning applied to document recognition.
\newblock \emph{Proceedings of the IEEE}, 86\penalty0 (11):\penalty0
  2278--2324, 1998.

\bibitem[Long et~al.(2015)Long, Shelhamer, and Darrell]{2015Fully}
J.~Long, E.~Shelhamer, and T.~Darrell.
\newblock Fully convolutional networks for semantic segmentation.
\newblock \emph{IEEE Transactions on Pattern Analysis and Machine
  Intelligence}, 39\penalty0 (4):\penalty0 640--651, 2015.

\bibitem[Ronneberger et~al.(2015)Ronneberger, Fischer, and
  Brox]{ronneberger2015u}
Olaf Ronneberger, Philipp Fischer, and Thomas Brox.
\newblock U-net: Convolutional networks for biomedical image segmentation.
\newblock In \emph{International Conference on Medical image computing and
  computer-assisted intervention}, pages 234--241. Springer, 2015.

\bibitem[Badrinarayanan et~al.(2017)Badrinarayanan, Kendall, and
  Cipolla]{badrinarayanan2017segnet}
Vijay Badrinarayanan, Alex Kendall, and Roberto Cipolla.
\newblock Segnet: A deep convolutional encoder-decoder architecture for image
  segmentation.
\newblock \emph{IEEE transactions on pattern analysis and machine
  intelligence}, 39\penalty0 (12):\penalty0 2481--2495, 2017.

\bibitem[Zhao et~al.(2017)Zhao, Shi, Qi, Wang, and Jia]{zhao2017pyramid}
Hengshuang Zhao, Jianping Shi, Xiaojuan Qi, Xiaogang Wang, and Jiaya Jia.
\newblock Pyramid scene parsing network.
\newblock In \emph{Proceedings of the IEEE conference on computer vision and
  pattern recognition}, pages 2881--2890, 2017.

\bibitem[Fu et~al.(2019)Fu, Liu, Tian, Li, Bao, Fang, and Lu]{fu2019dual}
Jun Fu, Jing Liu, Haijie Tian, Yong Li, Yongjun Bao, Zhiwei Fang, and Hanqing
  Lu.
\newblock Dual attention network for scene segmentation.
\newblock In \emph{Proceedings of the IEEE/CVF Conference on Computer Vision
  and Pattern Recognition}, pages 3146--3154, 2019.

\bibitem[Shaban et~al.(2017)Shaban, Bansal, Liu, Essa, and
  Boots]{shaban2017one}
Amirreza Shaban, Shray Bansal, Zhen Liu, Irfan Essa, and Byron Boots.
\newblock One-shot learning for semantic segmentation.
\newblock In \emph{proceedings of the British Machine Vision Conference}, pages
  6230--6239, 2017.

\bibitem[Zhang et~al.(2019)Zhang, Lin, Liu, Yao, and Shen]{zhang2019canet}
Chi Zhang, Guosheng Lin, Fayao Liu, Rui Yao, and Chunhua Shen.
\newblock Canet: Class-agnostic segmentation networks with iterative refinement
  and attentive few-shot learning.
\newblock In \emph{Proceedings of the IEEE/CVF Conference on Computer Vision
  and Pattern Recognition}, pages 5217--5226, 2019.

\bibitem[Wang et~al.(2019)Wang, Liew, Zou, Zhou, and Feng]{wang2019panet}
Kaixin Wang, Jun~Hao Liew, Yingtian Zou, Daquan Zhou, and Jiashi Feng.
\newblock Panet: Few-shot image semantic segmentation with prototype alignment.
\newblock In \emph{Proceedings of the IEEE/CVF International Conference on
  Computer Vision}, pages 9197--9206, 2019.

\bibitem[Tian et~al.(2020)Tian, Zhao, Shu, Yang, Li, and Jia]{tian2020prior}
Zhuotao Tian, Hengshuang Zhao, Michelle Shu, Zhicheng Yang, Ruiyu Li, and Jiaya
  Jia.
\newblock Prior guided feature enrichment network for few-shot segmentation.
\newblock \emph{IEEE Transactions on Pattern Analysis \& Machine Intelligence},
  2020.

\bibitem[Yang et~al.(2020{\natexlab{a}})Yang, Meng, Li, Wu, Xu, and
  Chen]{yang2020new}
Yuwei Yang, Fanman Meng, Hongliang Li, Qingbo Wu, Xiaolong Xu, and Shuai Chen.
\newblock A new local transformation module for few-shot segmentation.
\newblock In \emph{International Conference on Multimedia Modeling}, pages
  76--87. Springer, 2020{\natexlab{a}}.

\bibitem[Liu et~al.(2020{\natexlab{a}})Liu, Zhang, Lin, and Liu]{liu2020crnet}
Weide Liu, Chi Zhang, Guosheng Lin, and Fayao Liu.
\newblock Crnet: Cross-reference networks for few-shot segmentation.
\newblock In \emph{Proceedings of the IEEE/CVF Conference on Computer Vision
  and Pattern Recognition}, pages 4165--4173, 2020{\natexlab{a}}.

\bibitem[Zhou et~al.(2016)Zhou, Khosla, Lapedriza, Oliva, and
  Torralba]{zhou2016learning}
Bolei Zhou, Aditya Khosla, Agata Lapedriza, Aude Oliva, and Antonio Torralba.
\newblock Learning deep features for discriminative localization.
\newblock In \emph{Proceedings of the IEEE conference on computer vision and
  pattern recognition}, pages 2921--2929, 2016.

\bibitem[He et~al.(2017)He, Gkioxari, Doll{\'a}r, and Girshick]{he2017mask}
Kaiming He, Georgia Gkioxari, Piotr Doll{\'a}r, and Ross Girshick.
\newblock Mask r-cnn.
\newblock In \emph{Proceedings of the IEEE international conference on computer
  vision}, pages 2961--2969, 2017.

\bibitem[Hariharan et~al.(2014)Hariharan, Arbel{\'a}ez, Girshick, and
  Malik]{hariharan2014simultaneous}
Bharath Hariharan, Pablo Arbel{\'a}ez, Ross Girshick, and Jitendra Malik.
\newblock Simultaneous detection and segmentation.
\newblock In \emph{European conference on computer vision}, pages 297--312.
  Springer, 2014.

\bibitem[Kirillov et~al.(2019)Kirillov, He, Girshick, Rother, and
  Doll{\'a}r]{kirillov2019panoptic}
Alexander Kirillov, Kaiming He, Ross Girshick, Carsten Rother, and Piotr
  Doll{\'a}r.
\newblock Panoptic segmentation.
\newblock In \emph{Proceedings of the IEEE/CVF Conference on Computer Vision
  and Pattern Recognition}, pages 9404--9413, 2019.

\bibitem[Cheng et~al.(2021)Cheng, Schwing, and Kirillov]{cheng2021per}
Bowen Cheng, Alex Schwing, and Alexander Kirillov.
\newblock Per-pixel classification is not all you need for semantic
  segmentation.
\newblock \emph{Advances in Neural Information Processing Systems}, 34, 2021.

\bibitem[Chen et~al.(2018)Chen, Zhu, Papandreou, Schroff, and
  Adam]{chen2018encoder}
Liang-Chieh Chen, Yukun Zhu, George Papandreou, Florian Schroff, and Hartwig
  Adam.
\newblock Encoder-decoder with atrous separable convolution for semantic image
  segmentation.
\newblock In \emph{Proceedings of the European conference on computer vision
  (ECCV)}, pages 801--818, 2018.

\bibitem[Noh et~al.(2015)Noh, Hong, and Han]{noh2015learning}
Hyeonwoo Noh, Seunghoon Hong, and Bohyung Han.
\newblock Learning deconvolution network for semantic segmentation.
\newblock In \emph{Proceedings of the IEEE international conference on computer
  vision}, pages 1520--1528, 2015.

\bibitem[Dong and Xing(2018)]{dong2018few}
Nanqing Dong and Eric~P Xing.
\newblock Few-shot semantic segmentation with prototype learning.
\newblock In \emph{proceedings of the British Machine Vision Conference},
  volume~3, 2018.

\bibitem[Wang et~al.(2020{\natexlab{a}})Wang, Kong, Shen, Jiang, and
  Li]{wang2020solo}
Xinlong Wang, Tao Kong, Chunhua Shen, Yuning Jiang, and Lei Li.
\newblock Solo: Segmenting objects by locations.
\newblock In \emph{European Conference on Computer Vision}, pages 649--665.
  Springer, 2020{\natexlab{a}}.

\bibitem[Carion et~al.(2020)Carion, Massa, Synnaeve, Usunier, Kirillov, and
  Zagoruyko]{carion2020end}
Nicolas Carion, Francisco Massa, Gabriel Synnaeve, Nicolas Usunier, Alexander
  Kirillov, and Sergey Zagoruyko.
\newblock End-to-end object detection with transformers.
\newblock In \emph{European Conference on Computer Vision}, pages 213--229.
  Springer, 2020.

\bibitem[He et~al.(2016)He, Zhang, Ren, and Sun]{he2016deep}
Kaiming He, Xiangyu Zhang, Shaoqing Ren, and Jian Sun.
\newblock Deep residual learning for image recognition.
\newblock In \emph{Proceedings of the IEEE conference on computer vision and
  pattern recognition}, pages 770--778, 2016.

\bibitem[Krizhevsky et~al.(2012)Krizhevsky, Sutskever, and
  Hinton]{krizhevsky2012imagenet}
Alex Krizhevsky, Ilya Sutskever, and Geoffrey~E Hinton.
\newblock Imagenet classification with deep convolutional neural networks.
\newblock \emph{Advances in neural information processing systems},
  25:\penalty0 1097--1105, 2012.

\bibitem[Redmon et~al.(2016)Redmon, Divvala, Girshick, and
  Farhadi]{redmon2016you}
Joseph Redmon, Santosh Divvala, Ross Girshick, and Ali Farhadi.
\newblock You only look once: Unified, real-time object detection.
\newblock In \emph{Proceedings of the IEEE conference on computer vision and
  pattern recognition}, pages 779--788, 2016.

\bibitem[Liu et~al.(2018)Liu, Lehman, Molino, Petroski~Such, Frank, Sergeev,
  and Yosinski]{liu2018intriguing}
Rosanne Liu, Joel Lehman, Piero Molino, Felipe Petroski~Such, Eric Frank, Alex
  Sergeev, and Jason Yosinski.
\newblock An intriguing failing of convolutional neural networks and the
  coordconv solution.
\newblock \emph{Advances in Neural Information Processing Systems}, 31, 2018.

\bibitem[Everingham et~al.(2010)Everingham, Van~Gool, Williams, Winn, and
  Zisserman]{everingham2010pascal}
Mark Everingham, Luc Van~Gool, Christopher~KI Williams, John Winn, and Andrew
  Zisserman.
\newblock The pascal visual object classes (voc) challenge.
\newblock \emph{International journal of computer vision}, 88\penalty0
  (2):\penalty0 303--338, 2010.

\bibitem[Lin et~al.(2014)Lin, Maire, Belongie, Hays, Perona, Ramanan,
  Doll{\'a}r, and Zitnick]{lin2014microsoft}
Tsung-Yi Lin, Michael Maire, Serge Belongie, James Hays, Pietro Perona, Deva
  Ramanan, Piotr Doll{\'a}r, and C~Lawrence Zitnick.
\newblock Microsoft coco: Common objects in context.
\newblock In \emph{European conference on computer vision}, pages 740--755.
  Springer, 2014.

\bibitem[Chen et~al.(2017)Chen, Papandreou, Kokkinos, Murphy, and
  Yuille]{chen2017deeplab}
Liang-Chieh Chen, George Papandreou, Iasonas Kokkinos, Kevin Murphy, and Alan~L
  Yuille.
\newblock Deeplab: Semantic image segmentation with deep convolutional nets,
  atrous convolution, and fully connected crfs.
\newblock \emph{IEEE transactions on pattern analysis and machine
  intelligence}, 40\penalty0 (4):\penalty0 834--848, 2017.

\bibitem[Boudiaf et~al.(2021)Boudiaf, Kervadec, Masud, Piantanida, Ben~Ayed,
  and Dolz]{boudiaf2021few}
Malik Boudiaf, Hoel Kervadec, Ziko~Imtiaz Masud, Pablo Piantanida, Ismail
  Ben~Ayed, and Jose Dolz.
\newblock Few-shot segmentation without meta-learning: A good transductive
  inference is all you need?
\newblock In \emph{Proceedings of the IEEE/CVF Conference on Computer Vision
  and Pattern Recognition}, pages 13979--13988, 2021.

\bibitem[Zhang et~al.(2021)Zhang, Xiao, and Qin]{zhang2021self}
Bingfeng Zhang, Jimin Xiao, and Terry Qin.
\newblock Self-guided and cross-guided learning for few-shot segmentation.
\newblock In \emph{Proceedings of the IEEE/CVF Conference on Computer Vision
  and Pattern Recognition}, pages 8312--8321, 2021.

\bibitem[Liu et~al.(2020{\natexlab{b}})Liu, Zhang, Zhang, and He]{liu2020part}
Yongfei Liu, Xiangyi Zhang, Songyang Zhang, and Xuming He.
\newblock Part-aware prototype network for few-shot semantic segmentation.
\newblock In \emph{European Conference on Computer Vision}, pages 142--158.
  Springer, 2020{\natexlab{b}}.

\bibitem[Nguyen and Todorovic(2019)]{nguyen2019feature}
Khoi Nguyen and Sinisa Todorovic.
\newblock Feature weighting and boosting for few-shot segmentation.
\newblock In \emph{Proceedings of the IEEE/CVF International Conference on
  Computer Vision}, pages 622--631, 2019.

\bibitem[Wang et~al.(2020{\natexlab{b}})Wang, Zhang, Hu, Yang, Cao, and
  Zhen]{wang2020few}
Haochen Wang, Xudong Zhang, Yutao Hu, Yandan Yang, Xianbin Cao, and Xiantong
  Zhen.
\newblock Few-shot semantic segmentation with democratic attention networks.
\newblock In \emph{European Conference on Computer Vision}, pages 730--746.
  Springer, 2020{\natexlab{b}}.

\bibitem[Yang et~al.(2020{\natexlab{b}})Yang, Liu, Li, Jiao, and
  Ye]{yang2020prototype}
Boyu Yang, Chang Liu, Bohao Li, Jianbin Jiao, and Qixiang Ye.
\newblock Prototype mixture models for few-shot semantic segmentation.
\newblock In \emph{European Conference on Computer Vision}, pages 763--778.
  Springer, 2020{\natexlab{b}}.

\end{thebibliography}






\end{document}